\documentclass[
]{ceurart}
\usepackage{listings}
\usepackage[utf8]{inputenc}
\usepackage[T1]{fontenc}
\begin{document}

\copyrightyear{2025}
\copyrightclause{Challenge and Workshop (BC9): Large Language Models for Clinical and Biomedical NLP, International Joint Conference on Artificial Intelligence (IJCAI), August 16--22, 2025, Montreal, Canada.}

\conference{International Joint Conference on Artificial Intelligence (IJCAI), August 16--22, 2025, Montreal, Canada}

\title{CaresAI at BioCreative IX Track 1 - LLM for Biomedical QA}

\author[1,2]{Reem Abdel-Salam}[%
  email=reem.abdelsalam13@gmail.com,
]
\cormark[1]
\fnmark[1]

\author[3,2]{Mary Adewunmi}[%
  email=mary.adewunmi@menzies.edu.au,
]
\fnmark[1]

\author[4,2]{Modinat A. Abayomi}[%
  email=modinat.abayomi@bc.edu,
]
\fnmark[1]

\address[1]{Cairo University, Faculty of Engineering, Computer Engineering Department}
\address[2]{CaresAI, Australia}
\address[3]{Menzies School of Health Research, Charles Darwin University, NT, Australia}
\address[4]{Department of Biology, Boston College, Massachusetts, USA}

\cortext[1]{Corresponding author.}

\begin{abstract}
Large language models (LLMs) are increasingly evident for accurate question answering across various domains. However, rigorous evaluation of its performance on complex question-answering (QA) capabilities is essential before its deployment in real-world biomedical and healthcare applications. This paper presents our approach to the MedHopQA track of the BioCreative IX shared task, which focuses on multi-hop biomedical question answering involving diseases, genes, and chemicals. We adopt a supervised fine-tuning strategy leveraging LLaMA 3 8B, enhanced with a curated biomedical question–answer dataset compiled from external sources including BioASQ, MedQuAD, and TREC. Three experimental setups are explored: fine-tuning on combined short and long answers, short answers only, and long answers only. While our models demonstrate strong domain understanding, achieving concept-level accuracy scores of up to 0.8, their Exact Match (EM) scores remain significantly lower, particularly in the test phase. We introduce a two-stage inference pipeline for precise short-answer extraction to mitigate verbosity and improve alignment with evaluation metrics. Despite partial improvements, challenges persist in generating strictly formatted outputs. Our findings highlight the gap between semantic understanding and exact answer evaluation in biomedical LLM applications, motivating further research in output control and post-processing strategies.

\end{abstract}

\begin{keywords}
  LLM \sep
  Biomedical \sep
  QA \sep
  supervised finetuning \sep
  prompting \sep
  LLM \sep
  exact answer \sep
  \end{keywords}

\maketitle

\section{Introduction}

Recent advances in the scalability of transformers in large language models (LLMs) have demonstrated remarkable capabilities in question answering (QA) across diverse domains \cite{lin2025explore, ashish2017attention}. These models—trained on vast corpora and fine-tuned for various tasks—hold significant promise for transforming information retrieval and reasoning in specialised areas in biomedical and healthcare applications across medical exams, consumer health, and biomedical research \cite{singhal2025toward}. However, due to the critical and high-stakes nature of these domains, rigorous and domain-specific evaluation of LLM performance to tackle multihop reasoning remains a prerequisite before real-world deployment. It often involves complex reasoning such as over highly specialised knowledge, terminology, and interrelated concepts spanning diseases, genes, chemicals, and treatments. Conventional benchmarks such as BioASQ, PubMedQA, and MedQA have been instrumental in evaluating LLMs in alleviating these challenges. However, many of these datasets rely on extractive or single-hop QA formats that may not fully capture the reasoning complexity required in real-world biomedical settings.

To address these limitations, the BioCreative IX workshop introduced the MedHopQA \cite{MedHopQAoverview} shared task as part of its Track 1 initiative. The MedHopQA task is designed to evaluate the multi-step reasoning capabilities of LLMs through a curated set of 1,000 biomedical questions centered on diseases, genes, and chemicals, with a particular emphasis on rare diseases \cite{LuRN43961}. Unlike prior benchmarks, MedHopQA questions require reasoning across multiple documents and sources, thus presenting a more realistic and challenging QA scenario. The task encourages participants to build systems capable of interpreting and integrating information from various biomedical contexts. While the development set includes only 45 annotated questions with both short and long answers, a larger question set of 10,000, of which 1,000 are hidden and used for official evaluation through the leaderboard. This setup ensures a robust and unbiased assessment of system performance.

In this paper, we present our approach to the MedHopQA task using supervised fine-tuning of a pre-trained LLM, LLaMA 3 8B, supplemented with external biomedical QA datasets. Our work explores different training strategies, including fine-tuning on short, long, and combined answer formats. We also address key challenges related to answer precision and brevity through a post-processing pipeline aimed at extracting concise responses from verbose model outputs.

Through this work, our study advances the effectiveness of LLM in the context of multi-hop biomedical QA and highlights the limitations of verbosity and inconsistency that persist in generating accurately formatted answers under stringent evaluation metrics.

\section{Background and Related Work}
There has been enormous progress in previous biomedical QA models. Generally, these models can be categorised based on the type of task and the complexity of the reasoning process of such tasks.
\subsection{Classification of QA models}
\begin{enumerate}
\item {Task-based QA models} are objective and can be classified as extractive and abstractive QA. Extractive QA \cite{DBLP:journals/corr/abs-2110-03142} generates answers to a question by extracting a direct span of text from a given dataset. This model has been used in various tasks such as text extraction, document retrieval, and biomedical document QA. It is evaluated using metrics, Exact Match (EM) and F1-score. Instances of such models are: BERT\cite{koroteev2021bert}, RoBERTa \cite{roberts2022overview}, DeBERTa\cite{he2020deberta}, Albert \cite{lan2019albert}, DistilBERT \cite{sanh2019distilbert}. In contrast, Abstractive QA \cite{fan2019eli5}generates its answer in natural language, not necessarily copying from the given dataset. This model has been used in various tasks such as text paraphrasing, summarisation or synthesising new information. It is evaluated with ROUGE, BLEU, METEOR, BERTScore, and human evaluation. Instances of such models are GPT, T5, LLaMA, PaLM, and BART. The answers generated from both models can be yes/no, multiple choice, and open-domain.

\item Reasoning-based QA models, otherwise known as reasoning models, which provide answers depending on the complexity of the questions.
Single-Hop QA \cite{yuntao2022effective} provides answers to questions using a single source of information, which can be a sentence or a paragraph. Multi-Hop QA \cite{roy2022multi}involves providing an answer by combining multiple pieces of information from different parts or multiple documents.

Models can either fall into one or a combination of these classes, depending on the datasets used. For instance, BioASQ \cite{tsatsaronis2012bioasq} can be extractive, abstractive, single-hop, and multi-hop QA. MedQA \cite{yang2024llm} is largely abstractive, requiring multi-hop reasoning. BioGPT \cite{luo2022biogpt} is an abstractive QA and can perform both single-hop and multi-hop reasoning, depending on the prompt.
\end{enumerate}

\section{System Overview}
This section describes different stages for approaching shared tasks, including 1- Dataset, 2- Methodology and 3- Post-processing.

\subsection{Dataset}
The dataset provided by the shared task organisers comprises a development set and a test set \cite{LuRN43961}. The development set includes 45 question instances, each annotated with both a short and a long answer. These questions are primarily centered on biomedical topics, with a specific focus on diseases, genes, and chemicals, particularly those related to rare diseases. The dataset is curated from publicly available information on Wikipedia. The short answer typically consists of a single word or phrase, whereas the long answer presents a more comprehensive explanation intended to reflect the model’s underlying reasoning process. The test set contains 10,000 questions and serves as the primary benchmark for evaluating model performance in the shared task.
To effectively approach this task, we constructed a supplementary training dataset by aggregating biomedical question–answer pairs from various external sources. We utilized a diverse collection of biomedical and medical QA datasets, each varying in size and scope. Specifically, we included MedQuAD (16,412 QA pairs) \cite{BenAbacha-BMC-2019}, QALD (394) \cite{unger2014question}, MASH-QA (approximately 35,000) \cite{zhu-etal-2020-question}, MediQA (383) \cite{unger2014question}, Wikipedia Medical QA (1,000 manually collected entries), BiQA (8,216) \cite{krithara2023bioasq}, BioASQ (around 3,720) \cite{nentidis2023overview}, and TREC (5,500) \cite{roberts2022overview}.
To ensure quality and relevance, we applied a preprocessing step across all datasets to clean and filter out non-relevant or noisy question–answer pairs, retaining only medically meaningful and well-formed entries suitable for biomedical QA tasks.

For traning we have utilized 10,000 QA pairs from publicly available biomedical sources. We used the official 1,000-example development set released by the shared task organizers as our validation set. The official test set was reserved for final evaluation through the leaderboard and was not used during training.

\subsection{Methodology}

There are several strategies for addressing this task, including Retrieval-Augmented Generation (RAG), few-shot prompting, and supervised fine-tuning. In this work, we adopt the supervised fine-tuning approach, focusing on adapting a pre-trained large language model (LLM)—specifically, LLaMA 3 8B—to our curated biomedical dataset.

We conducted three experiments:
\begin{enumerate}
    \item Fine-tuning LLaMA 3 8B on the full dataset, including both short and long answers.
    \item Fine-tuning LLaMA 3 8B exclusively on short answer data.
    \item Fine-tuning LLaMA 3 8B exclusively on long answer data.
\end{enumerate}

Each of the three fine-tuned models was evaluated independently to explore the effect of answer format (short, long, or combined) on performance. We did not use a consensus or ensemble mechanism across models in the official submission. Instead, we analyzed the trade-offs of each training setup to understand their strengths: the short-answer fine-tuned model was better suited for precision, while the long-answer version captured broader context.

\subsubsection{Experimental Setup}

This section outlines the hyperparameter configurations, prompt templates, training infrastructure, and decoding strategies used in our experiments.

\paragraph{Hyperparameters}

The model was fine-tuned using 4-bit quantisation with LoRA adapters applied to both attention and linear layers. The LoRA configuration used a rank of $r=64$ and $\alpha=16$. We employed a learning rate of $1 \times 10^{-4}$, trained the model for 5 epochs, and utilized the \texttt{paged\_adamw\_8bit} optimiser along with a cosine learning rate scheduler. The random seed was set to 42 to ensure reproducibility.

\paragraph{Hardware Resources}

Training was conducted on two NVIDIA RTX 3080 Ti GPUs, with total training time approximately one day, while inference took an average 1 to 4 hours.

\paragraph{Prompting}

For the first and third experiments (combined and long answers), the following prompt template was used:

\begin{verbatim}
Question: {question} 
### Response##:
\end{verbatim}

For the second experiment (short answers), we used a domain-specific prompt designed to encourage concise biomedical responses:

\begin{verbatim}
You are a specialized biomedical doctor trained to answer clinical and biomedical questions. 
Your task is to generate a concise short answer that preserves
the full name of the entity in question.

Question: {question} 
### Response##:
\end{verbatim}

\paragraph{Decoding Strategies}
We applied a temperature of 0.01, top p: 0.95 for getting the answers from each model.

\subsection{Post-processing}
One of the key challenges encountered when fine-tuning and deploying the LLaMA 3 8B model across the three experimental setups was its tendency to produce overly verbose or imprecise answers. In many cases, the model did not generate a concise short answer, but instead responded by first restating the question, appending explanatory context, or extending the response with additional information beyond what was required. Table~\ref{tab:qa_examples} presents several examples of such outputs.

\begin{table}[h]
\centering
\caption{Examples of verbose or extended model outputs in response to short-answer questions.}
\begin{tabular}{|p{6cm}|p{7cm}|}
\hline
\textbf{Question} & \textbf{Model Output (Intended Short Answer)} \\
\hline
What type of protein aggregates are associated with Familial British dementia? & Amyloid-$\beta$ (A$\beta$) peptides are associated with Alzheimer's disease, but not with Familial British dementia. Instead, the primary aggregates are amyloid-related. \\
\hline
Which autoimmune disease is associated with chronic hives and can lead to dry mouth and dry eyes? & Sjögren's syndrome. Sjögren's syndrome is an autoimmune disease that primarily affects moisture-producing glands. \\
\hline
What is a common risk factor for both ectopic pregnancy and gastrointestinal diseases? & A common risk factor for both ectopic pregnancy and gastrointestinal diseases is pelvic inflammatory disease (PID). \\
\hline
\end{tabular}
\label{tab:qa_examples}
\end{table}

\begin{figure}[h]
    \centering
    \includegraphics[width=\linewidth]{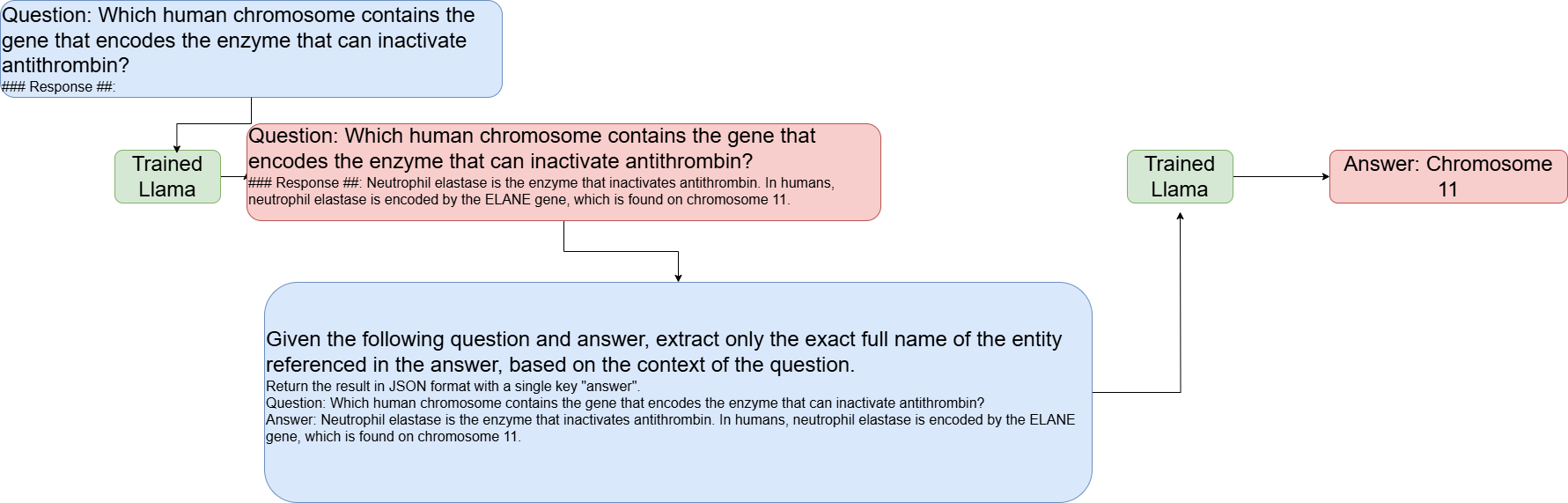}
    \caption{Two-stage inference pipeline for extracting precise short answers.}
    \label{fig:inference_pipeline}
\end{figure}

To address this issue, we designed a two-stage inference pipeline, illustrated in Figure~\ref{fig:inference_pipeline}. In the first stage, the model is prompted with the question to generate an initial response. In the second stage, a follow-up prompt is issued that explicitly instructs the model to extract the exact answer phrase or entity from the initial response. This post-processing stage is intended to ensure the final answer is succinct and aligned with the short-answer format.

However, this refinement process does not always yield a successful extraction. In cases where the model fails to produce a valid short answer across three sampled attempts, we default to using the longer generated response from the first stage as the final output. This fallback mechanism ensures that the system can still produce a semantically meaningful answer even when the model fails to generate a valid concise form after three attempts. While this does not always guarantee EM success, it maintains the usefulness of the output for downstream tasks. 

\section{Results}
\subsection{Evaluation Metrics}

Two primary metrics were used to evaluate model performance: Exact Match (EM) and concept-level evaluation. 

\paragraph{Exact Match (EM)} A prediction is considered correct under the EM metric if it exactly matches the gold standard answer after applying normalisation techniques. These include lowercasing, removal of punctuation and articles, and basic synonym resolution. This metric emphasises the precision of the model in generating strictly correct answer forms.

\paragraph{Concept-level Evaluation} This metric assesses whether the submitted answer is semantically equivalent to the gold answer, regardless of surface-level differences. It accounts for linguistic variation and aims to reflect the model's understanding of biomedical concepts.

\subsection{Final Results}

On the validation set, all three approaches achieved approximately 0.5 in EM score and around 0.8 in concept-level accuracy. These results suggest that the model has successfully captured biomedical knowledge and semantic understanding; however, it still faces challenges in producing exact matches, likely due to verbosity, paraphrasing, or formatting inconsistencies in its outputs.

In contrast, zero-shot inference using general-purpose models such as LLaMA 8B and Qwen Instruct 7B resulted in near-zero EM scores. This poor performance was mainly due to the generation of incorrect or irrelevant answers in most cases, and in some instances, a failure to produce concise, single-phrase answers as instructed. Upon closer inspection, LLaMA 8B often failed to follow the prompt to generate a short 1–2 phrase response, producing verbose or off-topic completions. Qwen 7B Instruct, on the other hand, demonstrated better adherence to the prompt format and instruction following.

Both models occasionally managed to answer surface-level factoid questions, such as "Which actor from the 1947 movie Driftwood, which features Rocky Mountain spotted fever, was born in Virginia City, Nevada?" However, they frequently hallucinated answers to questions requiring biomedical reasoning, especially those related to chromosomes, genes, or rare diseases. In contrast, questions asking about the appropriate medical field or specialty (e.g., "Which medical specialty treats...") were answered with approximately 80\% accuracy, indicating that while these models may capture some general domain patterns, they lack reliable biomedical knowledge and are prone to factual errors and hallucinations when dealing with specialized content.

In the testing phase, which was currently evaluated on 1,000 examples only, performance dropped significantly. The first approach yielded an EM score of 0.2, while the second and third approaches achieved an EM score of 0.0. This outcome highlights the model’s difficulty in generalizing to unseen data with precise answer formatting. It also underscores the need for further refinement in answer extraction strategies and post-processing methods, beyond simple sampling techniques.

Upon investigating the source of error, we found that a substantial portion of the performance drop was due to the model's inability to consistently generate concise 1–2 phrase answers, as required by the task instructions. In many cases, the model failed to format its output in the exact manner specified by the task organizers. For example, for a question like “On which chromosome is the gene located?”, the expected answer is “Chromosome 2”, whereas the model would often produce variants such as “2”, “Chr.2”, “2 chromosome”, or even extended spans like “Chromosome 2p13”. These deviations—despite containing correct content—lead to mismatches under strict exact match evaluation.

After addressing these issues through prompt refinement and lightweight post-processing in an unofficial test evaluation, the model was able to achieve a significantly improved EM score of 0.49.

These findings show that while the model understands biomedical concepts well, it still needs improvement in giving exact and correctly formatted answers, especially when strict evaluation rules are applied.

For concept-level results on the leaderboard, the first approach achieved a score of 0.3120, while the second and third approaches yielded 0.1140 and 0.2250, respectively.


\section{Conclusion}
In this work, we present a supervised fine-tuning approach using LLaMA 3 8B to tackle the MedHopQA shared task, which requires answering multihop biomedical questions with both short and long responses. We supplement the limited development data with external biomedical QA datasets and evaluate multiple fine-tuning strategies, including training on combined, short-only, and long-only answer sets. Our experiments demonstrate that while the model effectively captures biomedical semantics, achieving up to 0.8 in concept-level accuracy, it struggles to produce exact answer matches, with Exact Match (EM) scores plateauing around 0.5 on validation and dropping to as low as 0.0 on test data. This highlights the challenge of generating precise, well-formatted answers under strict evaluation constraints. To address the verbosity and inconsistency in short-answer generation, we implemented a two-stage inference pipeline aimed at extracting concise phrases from longer responses. Although this approach improved output quality in many cases, it did not consistently guarantee correct short answers. These findings underline the limitations of current large language models in tightly scoped answer generation and emphasise the need for more advanced post-processing, fine-tuning strategies, and evaluation-aligned prompt design. Future work will explore more robust extraction mechanisms and reinforcement learning techniques to align model outputs with evaluation criteria more reliably.

\bibliography{sample-ceur}


\end{document}